\pdfoutput=1

\documentclass[11pt]{article}

\usepackage{setspace}
\usepackage[dvipsnames]{xcolor}
\usepackage{graphicx}
\usepackage{tabularx}
\usepackage{afterpage}
\usepackage{booktabs}
\usepackage{amssymb,amsmath}
\usepackage{subcaption}
\usepackage{xspace}
\usepackage{etoolbox}
\usepackage{enumitem} %
\usepackage{comment} %

\usepackage{times,latexsym,xspace}
\usepackage[T1]{fontenc}
\usepackage[utf8]{inputenc}

\usepackage{microtype}

\usepackage{inconsolata}
\usepackage{float}
\usepackage{makecell}

\usepackage[final]{acl} %

\providetoggle{showcomments}
\toggletrue{showcomments}  %

\title{Adapting from Downturns: Prediction of Long-Term Conversational-Skill Development in Mental-Health Crisis Counselors
}

\author{Vivian Nguyen  \\ Cornell University \\  \texttt{vn72@cornell.edu} 
\And Lillian Lee \\ Cornell University \\  \texttt{llee@cs.cornell.edu}  \AND Elizabeth A.~Olson \\ Crisis Text Line \\ \texttt{eolson@crisistextline.org} \And Cristian Danescu-Niculescu-Mizil\thanks{Senior corresponding author.}  \\ Cornell University \\\texttt{cristian@cs.cornell.edu}}

\newcommand{\definedas}{\overset{{\rm def}}{=}}
\newcommand{\summ}[3]{\mathbf{z_{{#1}\rightarrow{#2}}}(#3)} 
\newcommand{\correctLabel}{y^*}

\newcommand{\disengagement}{disengagement\xspace}

\newcommand{\downturn}{downturn\xspace}

\newcommand{\downturns}{{\downturn}s\xspace}

\newcommand{\improve}{improve\xspace}

\definecolor{therapist}{RGB}{36,97,189}
\definecolor{client}{RGB}{244,130,130}

\newcommand{\cut}[1]{}
\newcommand{\xhdr}[1]{{\noindent\bfseries #1.}}

\makeatletter
\ifacl@anonymize
      \excludecomment{acksEnvironment}
\else
      \includecomment{acksEnvironment}
\fi
\makeatother

\begin{document}
\maketitle

\begin{abstract}

How do people learn to become better conversationalists?
This question is especially important in the context of mental-health counseling, where conversational skills are essential, yet volunteer counselors often have limited access to supervision and structured feedback. 
Understanding how counselors develop their ability to steer conversations toward positive outcomes---and identifying early which counselors are (not) on track to improve---can help prioritize support 
for the counselors who need it most.

In this work, we introduce the task of predicting, early in a conversationalist’s career, whether they will eventually improve at steering conversations toward positive outcomes, and demonstrate the feasibility of this task in the case of volunteer mental-health crisis counselors.
Our central insight is that people may struggle with particular kinds of moments in a conversation, and that what is especially revealing of their likelihood of future improvement is how they learn to handle those moments over time.
We operationalize this insight by designing a method that identifies 
the types of moments a counselor initially struggles with, 
captures how they adapt their response when they re-encounter similar moments in subsequent conversations, and learns which early adaptations predict improvement months or even years later.
While this future-prediction task is challenging, our counselor-adaptation approach yields better results than baselines 
that learn directly from the conversation transcript.

\end{abstract}

\section{Introduction and Related Work}
\label{sec:intro}

A large body of work has sought to understand why some conversations succeed while others fail. 
Computational work has modeled conversation-level outcomes \cite{lambert_conversational_2022}, identified behaviors associated with successful interactions \cite{althoff_large-scale_2016,niculae-danescu-niculescu-mizil-2016-conversational,bao_conversations_2021,de-kock-vlachos-2021-beg, shah_modeling_2022, yang_what_2024}, and developed methods for predicting when a conversation is likely to go off track 
\cite{perez-rosas_analyzing_2018,zhang_conversations_2018,chang_trouble_2019,kementchedjhieva_dynamic_2021,yuan_conversation_2023,imran_toxicity_2026}.
Much less is known about how conversationalists themselves improve over time: how they learn from repeated interactions to become better at steering conversations toward positive outcomes. 

Understanding this development process is especially important in high-stakes domains such as mental-health crisis counseling, where volunteer counselors regularly navigate challenging and extremely consequential conversations with limited access to supervision and feedback.
Within this setting, counselors vary substantially in their ability to steer 
conversations towards positive outcomes.
While some counselors 
improve
over time, increasing their success rate,
many improve only to a small degree or not at all,
as shown in \autoref{sec:improvement}
(a finding that holds for psychotherapists in general, per 
\citet{tracey_expertise_2014,goldberg_psychotherapists_2016,germer_does_2022}). 
When cases are randomly assigned (as is the case for the major crisis-counseling platform we work with), these differences reflect genuine distinctions in skill development rather than differences in case difficulty.

\begin{figure*}[t]
    \centering
    \includegraphics[width=1\linewidth]{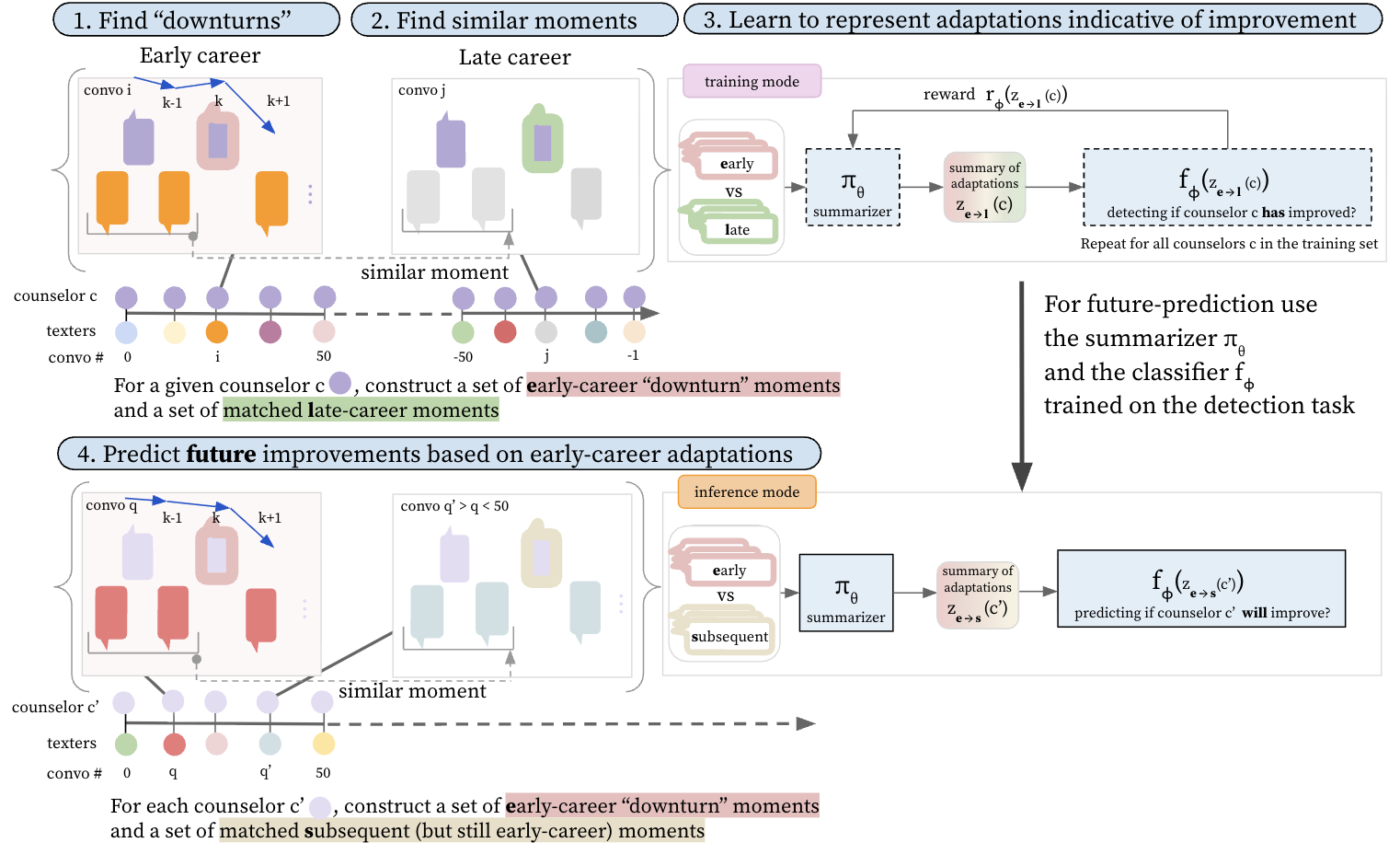}
    \caption{
    \label{fig:method-figure}
    High-level overview of our four-step approach.
    }
\end{figure*}

Being able to identify early on whether a counselor is likely to eventually improve (or not) can help prioritize support to those counselors who most need it.\footnote{
Provision of support for mental-health counselors is itself an important line of related work \cite{chaszczewicz-etal-2024-multi,louie-etal-2024-roleplay,srinivas-etal-2025-substance,yang-etal-2025-consistent,louie_simulated_2026,soma_artificial_2026}.
}
Hence, 
in this work we introduce the task of predicting whether volunteer counselors are on track to autonomously improve their ability to steer 
conversations towards positive outcomes.
Crucially, this prediction is made from conversations observed early in their activity as counselors (henceforth, their \textit{early career}), before their long-term trajectory is known.

Beyond the inherent difficulty of predicting the future, a key conceptual challenge is that there is no fixed playbook for becoming a better counselor:
we do not know a priori what improvement might look like. 
How can we, then, map a counselor's individual actions within a single conversation to long-term improvement?

\xhdr{High-level overview of our approach} 
We learn what improvement looks like empirically by first solving an easier post-hoc detection task: distinguishing the behavior of counselors who improved from those who did not (Steps~1--3 in \autoref{fig:method-figure}).
This teaches the model to recognize improvement-related changes in counselor behavior, which we refer to as \textit{adaptations}.
We then apply this learned representation to the early-prediction task: determining whether a counselor is \textit{already} on track to \improve in the future (Step~4).
We show that this model is able to extract predictive information from early-career signals that are otherwise difficult to learn from directly.

\xhdr{Contributions of Steps 1 \& 2: downturns}
While conversational behavior is highly complex, not all counselor actions are equally consequential.
Our insight is that improvement is especially visible in how counselors adapt to the types of moments they initially struggle with
\citep{dahlin_opportunity_2018,kapur_designing_2012}.
To identify such moments, we must connect conversation-level outcomes to the specific moments that most contributed to these outcomes. 
We do that by using a conversational forecasting model that is trained to predict the outcome after every utterance, allowing us to identify \textit{downturns}: moments where the predicted trajectory sharply worsens (Step~1).
We show that counselors experience different types of downturns early in their career, and that how they respond to similar situations in subsequent conversations (Step~2) is more predictive of long-term improvement than changes in their behavior elsewhere.

\xhdr{Contributions of Step 3: adaptation}
Finally, to use these insights, we must capture how counselors adapt their behavior across similar moments in different conversations. 
We represent these \textit{adaptations} as summaries of changes in conversational behavior, which we tune via reinforcement learning to focus on changes that are most indicative of improvement (Step~3).
This learned representation further improves  performance on the future-prediction task, demonstrating that it is feasible (albeit challenging) to predict early on whether a conversationalist will eventually 
\improve.

\vspace{0.1in}
\xhdr{Other domains} In this work, we focus on the mental-health crisis counseling domain, a particular type of high-stakes and asymmetric setting.
Whether the behavioral patterns associated with improvement in this setting extend to other conversational domains remains an open question.
Even so, our framework suggests a broader approach for studying how people develop conversational expertise over time, especially in domains where conversational skill is essential, such as education, coaching, and political debates. 
To encourage such new applications, we release our code together with a demonstration on public data 
as part of ConvoKit \cite{chang_convokit_2020}.\footnote{\url{https://convokit.cornell.edu/}}

\section{Data}
\label{sec:data}

In collaboration with the platform and with IRB approval, 
we study the conversational skill development of counselors on Crisis Text Line (CTL), a free, 24/7 crisis-counseling service that provides text-based messaging support for individuals ---henceforth \textit{texters}--- experiencing mental-health distress.
Texters come in with a wide range of challenges, from relationship issues to depression to suicidal ideation, 
and are randomly assigned to counselors (thus removing counselor selection as a possible confound).
The counselor's goal is to guide the texter to a calmer state. 

Volunteer counselors receive standardized training that includes extensive information about how to handle conversations, but does not provide explicit guidance as to how to improve over time. 
Two of the co-authors completed the training program to gain first-hand insights into the training process and the challenges counselors face.

The data contains over $1.5$M de-identified conversations from Jan 2015 to Oct 2020, with personally-identifiable information redacted by the platform. 
Many counselors in our data accumulate substantial time and activity on CTL:
$47$\% engage in more than $50$ conversations, and almost a quarter ($23$\%) hold more than $125$ conversations.
On average, counselors take $3$ months to reach $50$ conversations and $6$ months to reach $125$ conversations.

The data cannot be publicly shared due to its highly private nature, even in its redacted version.

\section{Setup:  Counselor Improvement}
\label{sec:improvement}
Here, we formalize conversational outcomes, quantify a counselor's ability to steer conversations towards positive outcomes, and define a measure of change in this ability across a counselor's career.
We combine these to formalize the task of predicting early on in a counselor's career whether, later in their career, they will 
get better at steering conversations towards positive outcomes 
(henceforth \textit{\improve}).

\xhdr{Measuring conversational outcomes} 
Assessing counseling conversational quality is known to be difficult as no single perfect measure exists \citep{tracey_expertise_2014, zhang_quantifying_2020, althoff_large-scale_2016, perez-rosas_analyzing_2018}.
We follow prior work in this domain and use conversational outcomes as a (precise but narrow) proxy for conversational quality: 
low-quality conversations are defined by \citet{nguyen_hanging_2025} as those that end with the texter \textit{disengaging} (leaving the conversation abruptly).
We extend their approach by validating and then using LLM-as-a-judge to evaluate how well a conversation ended on a $1$--$5$ scale, incorporating a predefined rubric and human-annotated examples.
Higher scores correspond to more positive conversational outcomes (e.g., texter feeling supported and calmer) and lower scores indicate undesirable outcomes (e.g., disengagement).
We find these labels to be in high agreement with human judgments ($94\%$).
Details and additional validation are provided in \autoref{appendix:convo_outcome}. %

\xhdr{Counselors' outcome tendencies}
Before studying improvement, we first ask whether conversational outcomes reflect counselor-specific tendencies, rather than pure randomness or short-term variation.
To assess this consistency, we take inspiration from \citet{zhang_quantifying_2020} and correlate counselors' average outcome scores across two interleaved sets of their conversations: their first 25 even-numbered conversations and their first 25 odd-numbered conversations. 
The significant positive correlation (Spearman's $\rho = 0.45, p < 0.0001$; compare with $\rho = 0.009, p =0.49$ when correlating outcomes from different counselors) suggests that the outcome score is not merely due to noise or short-term variation, but that it is a tendency specific to each counselor, and thus a meaningful basis for analyzing change over time.

\xhdr{Measuring improvement} 
To measure improvement, we compare counselor performance at two different points in their career.
We restrict our analysis to counselors with at least $125$ conversations (on average spending $20$ months on the platform) to allow sufficient opportunity for conversational skill development.
For each counselor, we define their first $50$ conversations as their \textit{early career} and the latest $50$ conversations as their \textit{late career}.
We then measure counselor performance in each time period by averaging the conversation outcome scores, resulting in $s_c^{\mathrm{early}}$ and $s_c^{\mathrm{late}}$ for counselor $c$.
We define a counselor's improvement score as the change in their average conversational outcome between their early and late career:
\begin{equation}
\Delta_c \definedas s_c^{\mathrm{late}} - s_c^{\mathrm{early}} \,.
\end{equation}
Using this measure, we find that, on average, counselors  improve their conversation outcomes from their early career to their late career ($p < 0.0001$; Wilcoxon signed-rank test). 
However, there is high variation in $\Delta_c$, with $36$\% of counselors not improving at all (see distribution in \autoref{fig:improvement} in the Appendix). 
We define \textit{improved} counselors as those whose improvement score $\Delta_i$ falls within the top quartile of all counselors, and \textit{non-improved} counselors as those within the bottom quartile.

Improved counselors show a reduction of variance in their outcomes, exhibiting more consistent outcomes in their late career than in their early career ($p < 0.0001$, Wilcoxon signed-rank test; distribution in \autoref{fig:change_variance} in the Appendix).
In contrast, non-improved counselors do not exhibit a comparable reduction in variance.
This suggests that counselor improvement is not primarily driven by a few isolated successes but instead corresponds to increasingly stable conversational skill.

\xhdr{Predicting future improvement}
Finally, we set up the task of predicting based on a counselor's early-career conversations whether they will improve in the future.
We use a matching procedure to control for baseline ability, allowing the experiment to focus specifically on forecasting future growth rather than distinguishing counselors who already differ substantially at the beginning of their careers (e.g., it is easier to improve from a low rate of positive outcome).
Specifically, for each improved counselor, we identify a non-improved counselor with the same (within a tolerance $t = 0.1$) early-career rate of positive outcomes.
This results in a balanced dataset of $1488$ counselors and $148800$ conversations (half early-career and half late-career) for our task, divided into a $65$/$15$/$20$ train/val/test split (dataset details summarized in \autoref{tab:dataset_stats}).

\begin{table}[t]
\centering
\small
\setlength{\tabcolsep}{5pt}
\begin{tabular}{p{0.65\columnwidth}p{0.28\columnwidth}}
\toprule
Statistic & Value \\
\midrule
Average conversation length (messages) & 28 \\
Average words per message & 22 \\
Average conversation duration (min) & 45 \\
Number of counselors & 1488 \\
Number of conversations & 148800 \\
\% conversations with scores 1, 2, 3, 4, 5 & 8, 20, 20, 29, 23  \\
\bottomrule
\end{tabular}
\caption{
Descriptive statistics of the dataset used for our improvement-prediction task.
}
\label{tab:dataset_stats}
\end{table}

\section{Method}
\label{sec:method}

As described in the Introduction and \autoref{fig:method-figure}, we first formalize and identify ``downturn'' moments where a counselor struggled (1), along with analogous moments later in their career (2).
Starting from changes in how counselors deal with these moments, we use a post-hoc detection task to learn to represent counselor adaptations in a way that is indicative of long-term improvement (3).
Finally, we apply this representation to capture early-career adaptations and predict whether a counselor is on track to improve in the future (4).

\subsection{Defining conversational \downturns}
\label{sec:define_downturn}
We aim to capture moments that counselors were challenged by.
We focus on conversations with undesirable endings (scores $\leq 2$), where counselors struggled to effectively navigate the interaction.
Although the undesirable outcome is only observed at the end of the conversation, our goal is to trace it back to the specific moments that might have been among the major contributors to that outcome.
To do so, we use a conversational forecasting model \cite{chang_trouble_2019, altarawneh_conversation_2023, kementchedjhieva_dynamic_2021}, which is trained to predict, as a conversation develops, its eventual outcome.
We use the method in prior work in this domain \cite{nguyen_hanging_2025} to train this model (details in \autoref{appendix:operationalization}). 
Formally, let $P(\textrm{\disengagement} \mid u_1 \ldots u_k)$ denote the forecasted probability of disengagement after observing the conversation up to turn $k$,
where $u_i$ is the content of turn $i$.
Intuitively, we consider a downturn to be a point where this forecasted probability dramatically worsens
(regardless of whether either the counselor or texter notices such a shift).

We formalize this notion of dramatic worsening using \textit{retrospective degradation} $RD$ (the reverse of retrospective improvement introduced by \citet{nguyen_hanging_2025}), which measures the increase in the probability of disengagement before and after a counselor turn $k$:

\begin{align*}
RD_{@k} \definedas P(\textrm{\disengagement}|u_1...u_{k+1}) - \\  P(\textrm{\disengagement} | u_1 \ldots u_{k-1}) \,.
\end{align*}
A high $RD$ indicates a moment where the trajectory of the conversation substantially worsened after the counselor's reply.

We then formalize a \textit{\downturn} in a conversation that ended poorly as the moment with  highest $RD$.
It is important to note that neither the counselor nor texter receives explicit signals about $RD$ values or \downturns either during or at the end of the conversation, and may never recognize any particular increase in $RD$.
Indeed, intuition suggests that a counselor who has difficulty recognizing moments of large $RD_{@k}$, either at turn $k+1$ or later on, might not easily learn to improve from experience (consistent with psychotherapy research on the importance of recognizing and repairing ruptures as a mechanism for change 
\citep{eubanks_introduction_2023}).

\xhdr{Alternatives to downturns}
In addition to downturns, formalized above, we could consider alternative moments to pinpoint and focus on.
One possibility is replacing downturns by a sentiment-based proxy, by selecting the moments where the sentiment worsens the most in the conversation (the most negative ``moment of change'', in the terminology of \citet{pruksachatkun_moments_2019}) rather than the probability of disengagement.
Another alternative is to simply choose moments randomly.\footnote{
Motivation Interviewing Skill Codes (MISC) annotations for characterizing client and counselor language \cite{houck_motivational_2010} could also be investigated, as could related annotation schemes, although developing accurate automatic labeling methods is an area of current research 
\cite{perez-rosas_analyzing_2018,singla_using_2018,cao-etal-2019-observing,shah_modeling_2022,mayer_predicting_2024}.
Two roadblocks in our setting are that we do not have ground-truth labels on our data to evaluate MISC-tagging accuracy, and, for privacy reasons, we cannot employ any of the LLMs that the most recent research applies (all models must be run internally).
}
Beyond the choice of moments, we could identify downturns across all conversations, instead of just conversations that ended unsuccessfully.
We will explore these alternatives to determine which moments yield representations more informative of counselor improvement.

\subsection{Learning representations of adaptations}
\label{sec:learning_adaptations}
Having identified moments a counselor was challenged by, we turn towards understanding how the counselor adapts to such moments over time.
Our goal is to learn a representation that captures how a counselor changes in these moments from their early to late career.
Importantly, we want to identify the behavioral changes most consequential for improvement, not simply how a counselor changes in general.
We focus on learning such adaptations specifically in downturns, and show that alternatives are less informative of improvement.

\xhdr{How counselors respond to downturns}
We aim to represent how counselors adapt to moments they previously struggled with.
For each counselor $c$, we compare how they initially responded in downturns with how they responded to analogous moments later on in their career. 
Concretely, for each downturn in a counselor's early career, we find the most contextually similar moment from the counselor's late career.
We consider the similarity between two moments as the cosine similarity of the last two texter messages before the counselor reply.\footnote{We validated the quality of the matches via a manual inspection of 50 matches.}

This aims to make the moments comparable, allowing us to isolate changes in the counselor's behavior from differences in context.
Aggregating across conversations yields two parallel sets of counselor responses for each counselor $c$:
\begin{itemize}
    \item \textbf{\underline{e}arly set}: a 
    counselor's
    responses at downturn
    moments in their early career
    \item \textbf{\underline{l}ate set}: a 
    counselor's
    responses at analogous moments in  their late career
\end{itemize}
We then prompt a large language model to summarize what changed between the early and late sets, producing a natural language description of the counselor's adaptations $\summ{e}{l}{c}$: how the counselor's behavior in these moments evolved from their \underline{e}arly $\rightarrow$ \underline{l}ate career.
We include prompts and implementation details in \autoref{appendix:operationalization}.

\begin{table}[t]
\centering
{\small
\setlength{\tabcolsep}{5pt}
\begin{tabular}{p{0.70\columnwidth}cc}
\toprule
Method & Val & Test \\ 
\midrule
\multicolumn{3}{l}{\textit{Downturn-based moments}} \\
\makecell[l]{Downturn (unsuccessful) \\
$\rightarrow$ paired similar moment} 
& \textbf{59.7} & \textbf{59.2} \\
\makecell[l]{Downturn (all) \\
$\rightarrow$ paired similar moment} 
& 59.3 & 56.9 \\
\makecell[l]{Downturn (unsuccessful) \\
$\rightarrow$ paired random moment} 
& 59.4 & 58.9 \\ \\
\multicolumn{3}{l}{\textit{Alternative moments}} \\
\makecell[l]{High sentiment drop (unsuccessful) \\
$\rightarrow$ paired similar moment} 
& 56.8 & 54.6 \\
\makecell[l]{Random moment (unsuccessful) \\
$\rightarrow$ paired similar moment} 
& 57.9 & 55.7 \\
\makecell[l]{Random moment (all) \\
$\rightarrow$ paired similar moment} 
& 55.2 & 52.2 \\
\midrule
\makecell[l]{Downturn (unsuccessful) \\
$\rightarrow$ paired similar moment \textbf{with RL}
} 
& 69.7 & 65.1 \\
\bottomrule
\end{tabular}
}
\caption{Accuracy in \textit{detecting} whether a counselor improved across different options for moment selections (early $\rightarrow$ late). The option that performed best on validation was used for reinforcement learning (bottom).}
\label{tab:detection_results}
\end{table}

\xhdr{Most informative moments}
We aim to have a representation that is indicative of counselor improvement.
We therefore explore multiple alternatives for moment selection and pairings on the validation set and pick the one that best captures this signal. 

To do so, we train an LLM classifier on these adaptations to detect improvement (training details in Appendix \ref{appendix:operationalization}).
\autoref{tab:detection_results} summarizes the results across different moment types. 
Across all types of moments, it is the use of downturns that results in representations of adaptations that are the most informative for detecting whether a counselor has improved;
the alternatives (random or sentiment-drop moments) produce substantially lower results.
Furthermore, selecting moments from all, rather than only unsuccessful, conversations lowers accuracy.
Therefore, we focus on adaptations from downturn moments moving forward.

\xhdr{Reinforcement-learned summarization of adaptations}
However, even in consequential moments, counselors change in many ways over the course of their careers, and not all of these changes are relevant to improvement.
A naive prompt-based summarizer may capture changes that are not meaningful for counselor development. 
For example, a summary describing a shift in a counselor's greeting style (e.g., ``hello'' to ``hi'') may not be relevant for determining whether a counselor has improved or not \cite{zhang_finding_2019}. 
Therefore, we need a way to steer the summarizer towards selecting what types of changes actually matter. 

In general, we want summaries of adaptations that are predictive of counselor improvement.
This requires training a summarizer LLM $\pi_\theta$ that produces (ideally informative but concise) descriptions of adaptations given a counselor $c$:
\begin{align*}
\summ{e}{l}{c} \sim \pi_\theta(\,\cdot \mid c)
\end{align*}
and a classifier
$f_\phi$ that uses the adaptation $\summ{e}{l}{c}$ to predict counselor improvement:
\begin{align*}
 \hat{y} = f_\phi(\summ{e}{l}{c})  \,.
\end{align*}
We propose a reinforcement-learning-based method that jointly trains both the summarizer $\pi_\theta$ and classifier $f_\phi$ for this task (\autoref{fig:method-figure}, Step 3).
The key intuition behind our approach is that an adaptation summary $\summ{e}{l}{c}$ is useful if it enables the classifier $f_\phi$ to accurately predict whether a counselor $c$ has improved.
Therefore, we design a reward function that rewards the summarizer based on the classifier's confidence in the correct label.
In other words, the summarizer receives higher reward when the classifier can use its adaptation summary to distinguish improved counselors from non-improved counselors.
Formally, we define the reward as the logit difference between the correct improvement label $\correctLabel \in \{0, 1\}$ for $c$ and incorrect label $1-\correctLabel$:
\begin{equation}
r_\phi(\summ{e}{l}{c}) = \text{logit}_\phi(\correctLabel) - \text{logit}_\phi(1-\correctLabel) \,.
\end{equation}
We use the logit difference rather than the classifier probability of the correct label, as probabilities near the decision boundary tend to be compressed by the softmax, flattening the reward landscape where the classifier is most uncertain. 
We find on the validation data that the logit difference preserves a stronger learning signal under this regime.

To jointly improve both models, we design an iterative training procedure, that alternates between improving the summarizer and improving the classifier.
We first initialize the summarizer $\pi_\theta$ and use its generated summaries to train the initial classifier $f_\phi$. 
Training then proceeds in an alternating manner where one component is updated at a time while the other is frozen. 
First, we optimize the summarizer $\pi_\theta$ via GRPO \cite{shao_deepseekmath_2024} using the reward $r_\phi(\summ{e}{l}{c})$ defined above.
Using the updated summarizer $\pi_\theta$, we then regenerate summaries and continue training $f_\phi$ on these newly generated summaries.
The updated classifier $f_\phi$ is used to continue to further optimize $\pi_\theta$.
This iterative method progressively steers the summarizer toward descriptions of adaptations relevant to improvement.   
At the same time, the classifier learns to better discriminate between improved and non-improved counselors based on these learned representations.

In training, we initialize this iterative procedure with the non-learned summaries that perform best on validation (i.e., those describing changes from downturn moments in unsuccessful conversations to analogous moments later; \autoref{tab:detection_results} top).
We pick the number of iterations by selecting the best performance on the validation set. There is extensive computational cost/time for training, as all work was done on secure internal servers due to the private nature of the data. The training time takes approximately 2 days on 3 x A6000 GPUs.
Training details are in \autoref{appendix:operationalization}.

As we will see, our procedure
leads to substantially better post-hoc improvement-detection results (\autoref{tab:detection_results} bottom).
We use both the improved representation and the improved classifier for solving our main future-prediction task, described next.

\subsection{Forecasting future improvement}
Finally we leverage the adaptation representation trained to detect improvement to identify whether a counselor's early-career changes already point toward future improvement (\autoref{fig:method-figure}, Step 4).
To do so, we adapt the contrastive setup to operate entirely within the early career. 
For each downturn $e$ in a counselor's \underline{e}arly career conversation, we retrieve the most contextually similar moment 
$s$ in a \underline{s}ubsequent conversation that is 
still within the early career.
We use the RL-trained summarizer $\pi_\theta$ trained on 
early $\rightarrow$ late changes
to generate a summary $\summ{e}{s}{\cdot}$ of this within-early-career change 
and apply the learned classifier $f_\phi$ to predict future improvement (for counselors in the test set).

The classifier $f_\phi$ was trained to detect improvement, discriminating between counselors who have already improved and those who have not. 
However, we are applying it to counselors in their early career to detect early signs of adaptation, before substantial long-term development has occurred.
To account for this distributional shift, we tune a threshold $\tau$ on the validation set, predicting a counselor $c'$ will improve if their score is above the threshold:

\begin{equation}
    \hat{y}_f = \mathbf{1}\bigl[f_\phi(\summ{e}{s}{c'}) \geq \tau\bigr].
\end{equation}
This allows us to capture whether a counselor appears to be moving in the right direction early in their career, before more substantial long-term improvement takes place.\footnote{
And indeed, the tuned threshold $\tau = 0.33$ is substantially lower than the $0.5$ threshold used for post-hoc detection, reflecting that the changes are less pronounced early on.}

\section{Results}
\label{sec:results}
\begin{table}[t]
\centering
\small
\begin{tabular}{lllcccc}
\toprule
Repres. & Train & RL & Acc & P & R & F1 \\ 
\midrule

\(\forall\) msg. 
& \(e\)
& 
& 56.9 & 57.6 & 53.2 & 54.8 \\

\(\times\) msg.
& \(e\)
& 
& 57.5 & 58.5 & 52.3 & 54.5 \\

diversity
& \(e\)
& 
& 54.6 & 54.4 & 57.0 & 55.7 \\
 
time 
& \(e\)
& 
& 47.0 & 46.7 & 43.6 & 45.1 \\

residual
& \(e\)
& 
& 56.7 & 55.4 & \textbf{68.4} & 61.2 \\

\midrule

\(\searrow\) adapt. 
& \(e \rightarrow l\) 
& \checkmark 
& \textbf{61.9} & \textbf{61.5} & 66.0 & \textbf{63.1} \\

\midrule

\(\searrow\) adapt.
& \(e \rightarrow s\) 
& \checkmark 
& 57.1 & 58.7 & 51.1 & 53.5 \\

\(\searrow\) adapt. 
& \(e \rightarrow l\) 
& 
& 55.1 & 58.8 & 39.3 & 45.7 \\

\(\searrow\) adapt.
& \(e \rightarrow s\) 
& 
& 53.4 & 53.5 & 52.2 & 52.6 \\

\(\searrow\) msg. 
& \(e\)
& 
& 60.3 & 60.6 & 59.3 & 59.8 \\

\bottomrule
\end{tabular}
\caption{
Results for the task of predicting future improvement from early-career conversations. We compare baselines (top) against our full system (middle), and ablations of the full system (bottom).
}
\label{tab:forecasting_results}
\end{table}

We now show that a counselor's early adaptation in downturn moments already encodes signals about their future trajectory. 
We compare our approach against several baselines and ablations of individual components of our method, testing the extent to which focusing on downturn moments or modeling adaptations is necessary (\autoref{tab:forecasting_results}).

\xhdr{Transcript baselines}
We first consider whether training an LLM classifier\footnote{All classifiers use the same base model  (Appendix \ref{appendix:operationalization}).} on raw anonymized transcripts of a counselor's early conversations is sufficient. 
\textbf{\(\forall\)~msg} uses counselor's messages from all early conversations (randomly sampled to fit in the context window), while
\textbf{\(\times\)~msg} is restricted to conversations that ended unsuccessfully.

\xhdr{Linguistic diversity baseline}
Beyond static conversational behavior, we also consider how a counselor changes their language across conversations as a signal of future growth \cite{zhang_finding_2019}.
A counselor who is on track to improve might exhibit a greater linguistic \textbf{diversity} early in their career, experimenting with different approaches rather than relying on a fixed set of templated responses. 
Using the ConvoKit implementation \cite{chang_convokit_2020}, we compute a counselor's linguistic diversity as the average cross-entropy of their messages under a unigram LM trained on their early conversations. 

\xhdr{Time baseline} We also consider the \textbf{time} it takes for counselors to reach $50$ conversations (the end of their early career).  
Counselors who take longer to reach this point might learn less, either because they encounter similar situations much later, or because they may be less motivated overall.
To test this hypothesis, we use the duration of a counselor's early career to predict future improvement, which generalizes poorly.

\xhdr{Residual baseline}
In constructing our task, counselors are paired based on their initial success rate (average conversational outcome score in their early career) within a fixed margin.  
To isolate this \textbf{residual} difference, we report results on predicting improvement based on early success rate alone.

\xhdr{Downturn results}
Rows in \autoref{tab:forecasting_results} that start with \(\searrow\) show the performance of models that focus on the downturns within a counselor's early conversations.
Training directly on a counselor's messages in their downturn moments (\(\searrow\) \textbf{msg}) 
outperforms directly using all their messages,
suggesting that the signal for future improvement is concentrated in downturns.

Importantly, \textit{how} a  counselor adapts after encountering these downturns matters.
Applying our full framework trained to detect post-hoc improvement based on \underline{e}arly $\rightarrow$ \underline{l}ate adaptations to downturns (\(\searrow\) \textbf{adapt.}  \(e \rightarrow l\)) achieves the highest performance.\footnote{Regardless of the training regime, all modes are applied using only the early-career conversations on the test set.
Our full system systematically outperforms the best baseline (\(\times\) msg.) across multiple seeds, as indicated in \autoref{fig:seed_results} in \autoref{appendix:additional-results}.

} 
This suggests that a counselor's potential for growth is not only signaled by the moments they struggled with, but also by whether they start to learn to handle subsequent moments in a way that reflects long-term improvement.

Training directly on \underline{e}arly$\rightarrow$\underline{s}ubsequent adaptations reduces performance, showing the importance of first learning what long-term improvements look like. 
Removing the RL step altogether also reduces performance, revealing the importance of learning which adaptations are indicative of improvement.

\section{Qualitative Analysis}
\label{sec:qualitative}
To add interpretability to our method, we examine qualitatively the types of downturns counselors experience and their respective adaptations.
\begin{table*}[ht]
\centering
\small
\setlength{\tabcolsep}{6pt}
\begin{tabularx}{\textwidth}{
    >{\raggedright\arraybackslash}X
    >{\raggedright\arraybackslash}X
    >{\raggedright\arraybackslash}X
}
\toprule
\textbf{Early-career response} & \textbf{Late-career response} & \textbf{Adaptation summary (excerpt)} \\
\midrule

\textbf{T:} I’m feeling suicidal \newline
\textbf{C:} Hi, I'm 
$\langle$redacted$\rangle$. I'm here for you, tell me more. \newline
\textbf{T:} I've been wanting to die for a week and idk what to do \newline
\textbf{C: You're brave for being so open with me. Would you be comfortable telling me your first name?}
&
\textbf{T:} I am suicidal \newline
\textbf{C:} Hi, I hear you're feeling overwhelmed tonight. Can you tell me more? \newline
\textbf{T:} I feel alone and wanna die \newline
\textbf{C: I'm here today to support you. What's happened recently that made you feel this way?}
&
[...] demonstrates a consistent pattern of offering support and expressing concern for the individual’s well-being [...] leans more heavily on direct statements of care [...] the approach is generally more proactive in seeking clarification about the individual’s situation and needs [...] \\
\midrule

\textbf{T:} I want to die. \newline
\textbf{C:} Has anything happened to make you want to self-harm? \newline
\textbf{T:} I felt terrible and getting blamed everything \newline
\textbf{C: I am sorry for your loss hon, it can be rather nasty when people unjustly blame you. Have you felt like this for a while?} 
&
\textbf{T:} I'll be dead soon. I wish I had spared myself years of pain. \newline
\textbf{C:} You're upset you are going to die soon? \newline
\textbf{T:} I am upset that my life ended. I am essentially already dead. \newline
\textbf{C: I am sorry to hear that you are feeling this way, though do you still want to continue talking?}
&
While [in early early-career responses the counselor] frequently used terms of endearment like 
``hon'' and offered reassurance, [in late-career responses the counselor] opts for more direct and questioning statements [...] consistently inquisitive tone [...] more distanced interaction style [...] move[s] towards a more objective communication. \\
\bottomrule
\end{tabularx}
\caption{
Paraphrased downturn and later analogous moments with adaptation summary excerpts. 
}
\label{tab:qual_examples}
\end{table*}

\xhdr{Downturn moments}
While downturns are shaped by the entire prior conversation context, to simplify the analysis we focus here on the two texter messages immediately preceding the downturn. 
For a systematic exploration, we use a Bayesian distinguishing-word analysis \cite{monroe_fightin_2017} to compare phrases that most distinguish these texter messages from ones at random moments.

Just before a conversation downturn, texters may \textit{explicitly request support}, mentioning they ``need help please'' or ``need someone to talk to''.
They express \textit{overwhelming distress} with a ``lot going on,'' ``a lot of anxiety'', or feeling like ``giving up''.
In the extreme, texters may also express \textit{suicidal ideation} or thoughts of self-harm (e.g., ``want to die'', ``end my life''). 
In such moments, counselors are put on the spot since their response may have high impact.
In contrast, moments that are not downturns provide fewer opportunities for the counselor to change the course of the conversation, for example after a texter has already decided to leave the conversation (e.g., ``have to go,'' ``going to sleep'', or ``don't want to talk anymore''). 
It is also worth noting that not all counselors encounter the same types of downturns.
The full list of distinguishing phrases, a clustering analysis of downturns, and the per-counselor distribution is discussed in \autoref{appendix:additional-qual}.

\xhdr{Adaptations} 
How do counselors adapt to these types of moments? 
\autoref{tab:qual_examples} provides example adaptations to downturns for two counselors, one who eventually improved and one who did not (more examples for other types of downturns are shown in \autoref{tab:qual_examples_all}).
In the first example, early in their career, a counselor replied to a texter expressing suicidal intentions 
by mentioning generically how the texter was ``brave for being so open'' and requesting them to share their name; this moment was identified as a downturn in a conversation that ended unsuccessfully with the texter suddenly leaving the conversation.
Later in their career, in a similar moment, the same counselor opts for a more active and supportive role, moving to quickly understand the texter's specific situation.
This change in approach is echoed by the summary of that counselor's adaptations: they learned  to ``lean more heavily on direct statements of care'' and to be ``more proactive in seeking clarifications about the [texter's] situation'' (note that this summary also reflects changes in other moments).

However, not all changes reflect improvements, as exemplified by the second example, from a counselor who did not improve.
They actually seem to be more reassuring in their early career, but later opted for ``more direct and questioning statements,'' a more ``consistently inquisitive tone'' and ``objective communication'' style.
Learning to distinguish which types of adaptations are indicative of improvements is a key component of our method.

\section{Conclusion}
\label{sec:discussion}
In this work, we show that it is feasible (albeit challenging) to predict, early in a counselor's career, whether they will eventually improve their ability to steer conversations towards positive outcomes.
We introduce a methodological framework that operationalizes a ``learning-from-mistakes'' intuition, which includes steps for detecting moments in which the counselors initially struggle and representing how counselors learn to change the way they approach similar moments encountered subsequently. 
Our results demonstrate the feasibility of this approach rather than presenting a deployment-ready system. 
Before such a framework can be used to inform real decisions about training or supervision, additional work is needed to assess its reliability, such as assessing its performance across specific counselor subgroups, and testing stability across cohorts and time.
We envision that after improvement and extensive testing, crisis-counseling platforms, which are critically dependent on volunteers, could use such a framework to focus additional training and supervision resources on a key population: those who stay on the platform, but do not initially find their footing and are not on track to improve just through experience.

\section{Limitations}
\label{sec:limitations}

In this work, we introduce the task of predicting whether a conversationalist will improve in the future, and demonstrate its feasibility in the specific context of volunteer mental-health crisis counselors.
While our approach outperforms all baselines, the performance gains are modest. 
This reflects the inherent difficulty of predicting the future using early behavioral signals alone (months or even years in advance). 
Importantly, this level of performance is comparable to the first conversation-level forecasting task \cite{zhang_conversations_2018} where initial performance was similarly limited, but since then has improved with more architectural and modeling advances \cite{chang_trouble_2019,yuan_conversation_2023,zhang_forecasting_2025}. 
In our case, we show that our task appears to be challenging for modern LLMs when applied directly to de-identified transcripts, and the present results can be viewed as a first step toward a broader line of work in studying a conversationalist's skill development, opening the door to future research in this area.

The primary contribution of the work is introducing the task and demonstrating that early counselor adaptation from downturns is predictive of long-term improvement. 
Individual components of our method (e.g., reward design, RL optimization, matching procedure) 
can be strengthened by future advances, and we view this work as establishing the feasibility of this approach rather than presenting its final implementation.

While our methodology is not inherently tied to our domain, it remains an open question whether the behavioral patterns associated with improvement in this setting extend to other conversational domains.
Our framework suggests a broader approach for studying how people develop conversational expertise over time, and we release our code to encourage new applications in other settings.

We also formulate the task as one of early prediction wherein we have fixed the improvement timeframe, i.e., improvement specifically from a counselor's early career to their late career.
Future work could extend this setting to an online formulation, where predictions on whether a counselor will improve can be made continuously throughout their career, as the counselor develops.

Our results should not be interpreted as applying to full-time professional therapists in general, but only to the  counselors on the platform we worked with, many of whom are trained volunteers.
Nonetheless, volunteer crisis counselors are a key population because of inadequate supply of mental-health support given growing demand \cite{ballout_trauma_2025}.

Our experiments involve counselors participating in at least 125 conversations in order to distinguish between early and late career. 
This selection criterion does limit our findings to relatively long-term counselors. Retention of counselors is a crucial issue, but beyond the scope of our work.

As a feasibility demonstration, we formulated prediction tasks requiring distinguishing between the top and bottom quartile of improvement.  
In application, predicting the actual degree of improvement may be necessary.

We worked from a ``learning-from-mistakes'' perspective, experimenting with ``negative'' situations such as \downturns, drops in sentiment, and so on.  
Future work could consider other types of moments, such as ``upturns'' and moments of change.

As mentioned in \autoref{sec:improvement}, there is no single perfect measure of the quality of a counseling conversation.  
While one could argue that using texter disengagement as measure is narrow, it is particularly important in the context of crisis counseling where the goal is for the texter to reach a calmer state by the end of the conversation. 
When a texter leaves before this amelioration process is complete or continues to express serious distress, the outcome is negative. 
We note that in rare cases, disengagement could also stem from incidental causes unrelated to the conversation itself (e.g., a dead phone battery) which our measure may not distinguish from genuine dissatisfaction or distress.
Even so, focusing on disengagement, while necessarily coarse-grained and possibly missing other aspects of conversation quality, captures an important and clinically meaningful outcome as an incomplete conversation may leave the texter at high risk. 
Unlike ongoing therapy sessions, one-off crisis counseling conversations do not provide an opportunity for the relationship to evolve over subsequent sessions, making resolution within the conversation particularly important.

We use LLM-as-a-judge to label conversation outcomes, and validate it by comparing it with rule-based methods and human judgment.  
However, the number of human judgments is limited due to privacy restrictions on the data. 
Future work could involve using responses to texter follow-up surveys as one possible alternate quality signal if response rates are sufficiently high.

\section{Ethical considerations}
The data we work with is highly sensitive.  PII was redacted by the providing platform before the research began.  
The data was kept on internal servers with access restricted to the authors, and all language models we employed were trained locally; in particular, we never supplied any of the data as prompt or input to any external LLMs. 

Our work does \textit{not} represent a use of generative AI to replace human counselors \cite{imel_framework_2026}. 
We acknowledge the potential risks of AI use in psychological services \cite{van_zyl_unintended_2026}.

Our intent in this research is to help crisis-counseling platforms \emph{enhance} the conversational skills of volunteer counselors.
Predicting volunteer counselor improvement trajectories is only the first step. 
Future work could identify routes to support all volunteers across this spectrum of predicted improvement, to develop and support their skill sets. 
Optimizing their development will result in the strongest outcomes, both for texters in crisis and for the volunteers themselves, as they accrue benefits from the experience of volunteering.

\begin{acksEnvironment}
\paragraph{Acknowledgments}

We thank the anonymous (meta)reviewers for valuable feedback, Dana Atzil for helpful conversations, and Sean Zhang for the data used in the demo.
We are also grateful for repeated conversations with members of Team Zissou, including Christina Bai, Alex Chen, Laerdon Kim, Sophia Liu, Claire Okamoto,  Ethan Xia, and Sean Zhang. 
This research would not have been possible without the support of Crisis Text Line; we are particularly grateful to Margaret Meagher and Lili Török for insights on the dynamics of crisis counseling conversations and for valuable feedback on earlier drafts.
Cristian Danescu-Niculescu-Mizil was funded in part by the U.S. National Science Foundation under Grant No. IIS-1750615 (CAREER) and the Cornell Center for Social Sciences.
Elizabeth A. Olson was funded in part by support to Crisis Text Line from the Jensen and Lori Huang Foundation.
Vivian Nguyen was supported by the GEM Fellowship.
Any opinions, findings, and conclusions in this work are those of the author(s) and do not necessarily reflect the views of Cornell University or the National Science Foundation.

\end{acksEnvironment}
 
\bibliography{refs_fixed}

@article{lambert_conversational_2022,
	title = {Conversational {Resilience}: {Quantifying} and {Predicting} {Conversational} {Outcomes} {Following} {Adverse} {Events}},
	journal = {Proceedings of the International AAAI Conference on Web and Social Media},
	author = {Lambert, Charlotte and Rajagopal, Ananya and Chandrasekharan, Eshwar},
	year = {2022},
}

@article{althoff_large-scale_2016,
	title = {Large-scale {Analysis} of {Counseling} {Conversations}: {An} {Application} of {Natural} {Language} {Processing} to {Mental} {Health}},
	journal = {Transactions of the Association for Computational Linguistics},
	author = {Althoff, Tim and Clark, Kevin and Leskovec, Jure},
	year = {2016},
}

@inproceedings{niculae-danescu-niculescu-mizil-2016-conversational,
	title = {Conversational {Markers} of {Constructive} {Discussions}},
	booktitle = {Proceedings of {NAACL}},
	author = {Niculae, Vlad and Danescu-Niculescu-Mizil, Cristian},
	year = {2016},
}

@inproceedings{bao_conversations_2021,
	title = {Conversations {Gone} {Alright}: {Quantifying} and {Predicting} {Prosocial} {Outcomes} in {Online} {Conversations}},
	booktitle = {Proceedings of the {WWW}},
	author = {Bao, Jiajun and Wu, Junjie and Zhang, Yiming and Chandrasekharan, Eshwar and Jurgens, David},
	year = {2021},
}

@inproceedings{de-kock-vlachos-2021-beg,
	title = {I {Beg} to {Differ}: {A} study of constructive disagreement in online conversations},
	booktitle = {Proceedings of {EACL}},
	author = {De Kock, Christine and Vlachos, Andreas},
	year = {2021},
}

@article{shah_modeling_2022,
	title = {Modeling {Motivational} {Interviewing} {Strategies} on an {Online} {Peer}-to-{Peer} {Counseling} {Platform}},
	number = {CSCW2},
	journal = {Proc. ACM Hum.-Comput. Interact.},
	author = {Shah, Raj Sanjay and Holt, Faye and Hayati, Shirley Anugrah and Agarwal, Aastha and Wang, Yi-Chia and Kraut, Robert E. and Yang, Diyi},
	year = {2022},
}

@article{yang_what_2024,
	title = {What {Makes} {Digital} {Support} {Effective}? {How} {Therapeutic} {Skills} {Affect} {Clinical} {Well}-{Being}},
	number = {CSCW1},
	journal = {Proc. ACM Hum.-Comput. Interact.},
	author = {Yang, Wenjie and Fang, Anna and Shah, Raj Sanjay and Mathur, Yash and Yang, Diyi and Zhu, Haiyi and Kraut, Robert E.},
	year = {2024},
}

@inproceedings{perez-rosas_analyzing_2018,
	title = {Analyzing the {Quality} of {Counseling} {Conversations}: the {Tell}-{Tale} {Signs} of {High}-quality {Counseling}},
	booktitle = {Proceedings of {LREC}},
	author = {Pérez-Rosas, Verónica and Sun, Xuetong and Li, Christy and Wang, Yuchen and Resnicow, Kenneth and Mihalcea, Rada},
	year = {2018},
}

@inproceedings{zhang_conversations_2018,
	title = {Conversations {Gone} {Awry}: {Detecting} {Early} {Signs} of {Conversational} {Failure}},
	booktitle = {Proceedings of {ACL}},
	author = {Zhang, Justine and Chang, Jonathan and Danescu-Niculescu-Mizil, Cristian and Dixon, Lucas and Hua, Yiqing and Taraborelli, Dario and Thain, Nithum},
	year = {2018},
}

@inproceedings{chang_trouble_2019,
	title = {Trouble on the {Horizon}: {Forecasting} the {Derailment} of {Online} {Conversations} as they {Develop}},
	booktitle = {Proceedings of {EMNLP}},
	author = {Chang, Jonathan P. and Danescu-Niculescu-Mizil, Cristian},
	year = {2019},
}

@inproceedings{kementchedjhieva_dynamic_2021,
	title = {Dynamic {Forecasting} of {Conversation} {Derailment}},
	booktitle = {Proceedings of {EMNLP}},
	author = {Kementchedjhieva, Yova and Sogaard, Anders},
	year = {2021},
}

@article{yuan_conversation_2023,
	title = {Conversation {Modeling} to {Predict} {Derailment}},
	journal = {Proceedings of the International AAAI Conference on Web and Social Media},
	author = {Yuan, Jiaqing and Singh, Munindar P.},
	year = {2023},
}

@inproceedings{imran_toxicity_2026,
	title = {Toxicity {Ahead}: {Forecasting} {Conversational} {Derailment} on {GitHub}},
	booktitle = {Proceedings of the International Conference on Software Engineering},
	author = {Imran, Mia Mohammad and Zita, Robert and Rahman, Rahat Rizvi and Chatterjee, Preetha and Damevski, Kostadin},
	year = {2026},
}

@article{tracey_expertise_2014,
	title = {Expertise in psychotherapy: an elusive goal?},
	journal = {The American Psychologist},
	author = {Tracey, Terence J. G. and Wampold, Bruce E. and Lichtenberg, James W. and Goodyear, Rodney K.},
	year = {2014},
}

@article{goldberg_psychotherapists_2016,
	title = {Do psychotherapists improve with time and experience? {A} longitudinal analysis of outcomes in a clinical setting},
	journal = {Journal of Counseling Psychology},
	author = {Goldberg, Simon B. and Rousmaniere, Tony and Miller, Scott D. and Whipple, Jason and Nielsen, Stevan Lars and Hoyt, William T. and Wampold, Bruce E.},
	year = {2016},
}

@article{germer_does_2022,
	title = {Does practice really make perfect? {A} longitudinal analysis of the relationship between therapist experience and therapy outcome: {A} replication of {Goldberg}, {Rousmaniere}, et al. (2016)},
	journal = {Journal of Counseling Psychology},
	author = {Germer, Sylvan and Weyrich, Vanessa and Bräscher, Anne-Kathrin and Mütze, Kaline and Witthöft, Michael},
	year = {2022},
}

@inproceedings{chaszczewicz-etal-2024-multi,
	title = {Multi-{Level} {Feedback} {Generation} with {Large} {Language} {Models} for {Empowering} {Novice} {Peer} {Counselors}},
	booktitle = {Proceedings of {ACL}},
	author = {Chaszczewicz, Alicja and Shah, Raj and Louie, Ryan and Arnow, Bruce and Kraut, Robert and Yang, Diyi},
	year = {2024},
}

@inproceedings{louie-etal-2024-roleplay,
	title = {Roleplay-doh: {Enabling} {Domain}-experts to {Create} {LLM}-simulated {Patients} via {Eliciting} and {Adhering} to {Principles}},
	booktitle = {Proceedings of {EMNLP}},
	author = {Louie, Ryan and Nandi, Ananjan and Fang, William and Chang, Cheng and Brunskill, Emma and Yang, Diyi},
	year = {2024},
}

@inproceedings{srinivas-etal-2025-substance,
	title = {Substance over {Style}: {Evaluating} {Proactive} {Conversational} {Coaching} {Agents}},
	booktitle = {Proceedings of {ACL}},
	author = {Srinivas, Vidya and Xu, Xuhai and Liu, Xin and Ayush, Kumar and Galatzer-Levy, Isaac and Patel, Shwetak and McDuff, Daniel and Althoff, Tim},
	year = {2025},
}

@inproceedings{yang-etal-2025-consistent,
	title = {Consistent {Client} {Simulation} for {Motivational} {Interviewing}-based {Counseling}},
	booktitle = {Proceedings of {ACL}},
	author = {Yang, Yizhe and Achananuparp, Palakorn and Huang, Heyan and Jiang, Jing and Lim, Nicholas Gabriel and Ern, Cameron Tan Shi and Kit, Phey Ling and Xiuhui, Jenny Giam and Pinto, John and Lim, Ee-Peng},
	year = {2025},
}

@inproceedings{louie_simulated_2026,
	title = {Can {LLM}-{Simulated} {Practice} and {Feedback} {Upskill} {Human} {Counselors}? {A} {Randomized} {Study} with 90+ {Novice} {Counselors}},
	booktitle = {Proceedings of {CHI}},
	author = {Louie, Ryan and Shah, Raj Sanjay and Orney, Ifdita Hasan and Pacheco, Juan Pablo and Brunskill, Emma and Yang, Diyi},
	year = {2026},
}

@article{soma_artificial_2026,
	title = {Artificial {Intelligence} to {Support} {Human}-{Provided} {Mental} {Health} {Treatment}},
	journal = {Annual Review of Clinical Psychology},
	author = {Soma, Christina S. and Kuo, Patty B. and Mehta, Maitrey and Srikumar, Vivek and Imel, Zac E. and Atkins, David C.},
	year = {2026},
}

@inproceedings{nguyen_hanging_2025,
	title = {Hanging in the {Balance}: {Pivotal} {Moments} in {Crisis} {Counseling} {Conversations}},
	booktitle = {Proceedings of {ACL}},
	author = {Nguyen, Vivian and Lee, Lillian and Danescu-Niculescu-Mizil, Cristian},
	year = {2025},
}

@article{zhang_quantifying_2020,
	title = {Quantifying the {Causal} {Effects} of {Conversational} {Tendencies}},
	journal = {Proceedings of the ACM on Human-Computer Interaction},
	author = {Zhang, Justine and Mullainathan, Sendhil and Danescu-Niculescu-Mizil, Cristian},
	year = {2020},
}

@inproceedings{chang_convokit_2020,
	title = {{ConvoKit}: {A} {Toolkit} for the {Analysis} of {Conversations}},
	booktitle = {Proceedings of {SIGDIAL}},
	author = {Chang, Jonathan P. and Chiam, Caleb and Fu, Liye and Wang, Andrew and Zhang, Justine and Danescu-Niculescu-Mizil, Cristian},
	year = {2020},
}

@article{dahlin_opportunity_2018,
	title = {Opportunity, {Motivation}, and {Ability} to {Learn} from {Failures} and {Errors}: {Review}, {Synthesis}, and {Ways} to {Move} {Forward}},
	journal = {Academy of Management Annals},
	author = {Dahlin, Kristina B. and Chuang, You-Ta and Roulet, Thomas J.},
	year = {2018},
}

@article{kapur_designing_2012,
	title = {Designing for {Productive} {Failure}},
	journal = {Journal of the Learning Sciences},
	author = {Kapur, Manu and Bielaczyc, Katerine},
	year = {2012},
}

@inproceedings{altarawneh_conversation_2023,
	title = {Conversation {Derailment} {Forecasting} with {Graph} {Convolutional} {Networks}},
	booktitle = {Proceedings of the 7th {Workshop} on {Online} {Abuse} and {Harms}},
	author = {Altarawneh, Enas and Agrawal, Ameeta and Jenkin, Michael and Papagelis, Manos},
	year = {2023},
}

@incollection{eubanks_introduction_2023,
	title = {Introduction: {Rupture} in a wicked and wonderful world},
	booktitle = {Rupture and repair in psychotherapy: {A} critical process for change},
	publisher = {American Psychological Association},
	address = {Washington},
	author = {Muran, J. Christopher and Eubanks, Catherine F. and Samstag, Lisa Wallner},
	editor = {Eubanks, Catherine F. and Samstag, Lisa Wallner and Muran, J. Christopher},
	year = {2023},
}

@inproceedings{pruksachatkun_moments_2019,
	title = {Moments of {Change}: {Analyzing} {Peer}-{Based} {Cognitive} {Support} in {Online} {Mental} {Health} {Forums}},
	booktitle = {Proceedings of {CHI}},
	author = {Pruksachatkun, Yada and Pendse, Sachin R. and Sharma, Amit},
	year = {2019},
}

@misc{houck_motivational_2010,
	title = {Motivational {Interviewing} {Skill} {Code} ({MISC}) 2.5},
	howpublished = {https://casaa.unm.edu/assets/docs/misc25.pdf},
	author = {Houck, Jon M. and Moyers, Theresa B. and Miller, William R. and Glynn, Lisa H. and Hallgren, Kevin A.},
	year = {2010},
}

@inproceedings{singla_using_2018,
	title = {Using {Prosodic} and {Lexical} {Information} for {Learning} {Utterance}-level {Behaviors} in {Psychotherapy}},
	booktitle = {Interspeech 2018},
	author = {Singla, Karan and Chen, Zhuohao and Flemotomos, Nikolaos and Gibson, James and Can, Dogan and Atkins, David and Narayanan, Shrikanth},
	year = {2018},
}

@inproceedings{cao-etal-2019-observing,
	title = {Observing {Dialogue} in {Therapy}: {Categorizing} and {Forecasting} {Behavioral} {Codes}},
	booktitle = {Proceedings of {ACL}},
	author = {Cao, Jie and Tanana, Michael and Imel, Zac and Poitras, Eric and Atkins, David and Srikumar, Vivek},
	year = {2019},
}

@inproceedings{mayer_predicting_2024,
	title = {Predicting {Client} {Emotions} and {Therapist} {Interventions} in {Psychotherapy} {Dialogues}},
	booktitle = {Proceedings of {EACL}},
	author = {Mayer, Tobias and Warikoo, Neha and Eliassaf, Amir and Atzil-Slonim, Dana and Gurevych, Iryna},
	year = {2024},
}

@article{shao_deepseekmath_2024,
  title = {{DeepSeekMath}: {Pushing} the {Limits} of {Mathematical} {Reasoning} in {Open} {Language} {Models}},
  journal = {arXiv preprint arXiv:2402.03300},
  author = {Shao, Zhihong and Wang, Peiyi and Zhu, Qihao and Xu, Runxin and Song, Junxiao and Bi, Xiao and Zhang, Haowei and Zhang, Mingchuan and Li, Y. K. and Wu, Y. and Guo, Daya},
  year = {2024},
}

@inproceedings{zhang_finding_2019,
	title = {Finding {Your} {Voice}: {The} {Linguistic} {Development} of {Mental} {Health} {Counselors}},
	booktitle = {Proceedings of {ACL}},
	author = {Zhang, Justine and Filbin, Robert and Morrison, Christine and Weiser, Jaclyn and Danescu-Niculescu-Mizil, Cristian},
	year = {2019},
}

@inproceedings{hu_lora_2021,
	title = {{LoRA}: {Low}-{Rank} {Adaptation} of {Large} {Language} {Models}},
    booktitle = {Proceedings of {ICLR}},
	author = {Hu, Edward J. and Shen, Yelong and Wallis, Phillip and Allen-Zhu, Zeyuan and Li, Yuanzhi and Wang, Shean and Wang, Lu and Chen, Weizhu},
	year = {2022},
}

@inproceedings{loshchilov_decoupled_2017,
	title = {Decoupled {Weight} {Decay} {Regularization}},
	booktitle = {Proceedings of {ICLR}},
	author = {Loshchilov, Ilya and Hutter, Frank},
	year = {2019},
}

@inproceedings{hu-etal-2025-openrlhf,
	title = {{OpenRLHF}: {A} {Ray}-based {Easy}-to-use, {Scalable} and {High}-performance {RLHF} {Framework}},
	booktitle = {Proceedings of {EMNLP}},
	author = {Hu, Jian and Wu, Xibin and Shen, Wei and Liu, Jason Klein and Wang, Weixun and Jiang, Songlin and Wang, Haoran and Chen, Hao and Chen, Bin and Fang, Wenkai and Xianyu and Cao, Yu and Xu, Haotian and Liu, Yiming},
	year = {2025},
}

@misc{liu_roberta_2019,
	title = {{RoBERTa}: {A} {Robustly} {Optimized} {BERT} {Pretraining} {Approach}},
	author = {Liu, Yinhan and Ott, Myle and Goyal, Naman and Du, Jingfei and Joshi, Mandar and Chen, Danqi and Levy, Omer and Lewis, Mike and Zettlemoyer, Luke and Stoyanov, Veselin},
	year = {2019},
}

@article{riviere2024gemma2,
	title = {Gemma 2: {Improving} {Open} {Language} {Models} at a {Practical} {Size}},
	journal = {arXiv preprint arXiv:2408.00118},
	author = {Riviere, Morgane and Pathak, Shreya and others},
	year = {2024},
}

@article{gemmateam2025gemma3,
	title = {Gemma 3 {Technical} {Report}},
	journal = {arXiv preprint arXiv:2503.19786},
	author = {Kamath, Aishwarya and Ferret, Johan and Pathak, Shreya and Vieillard, Nino and others},
	year = {2025},
}

@article{monroe_fightin_2017,
	title = {Fightin' {Words}: {Lexical} {Feature} {Selection} and {Evaluation} for {Identifying} the {Content} of {Political} {Conflict}},
	journal = {Political Analysis},
	author = {Monroe, Burt L. and Colaresi, Michael P. and Quinn, Kevin M.},
	year = {2008},
}

@article{imel_framework_2026,
	title = {A {Framework} for {Automation} in {Psychotherapy}},
	journal = {Current Directions in Psychological Science},
	author = {Imel, Zac E. and Creed, Torrey and Kious, Brent and Althoff, Tim and Atzil-Slonim, Dana and Srikumar, Vivek},
	year = {2026},
}

@article{van_zyl_unintended_2026,
	title = {The unintended negative consequences of artificial intelligence use for psychologists},
	journal = {Frontiers in Psychology},
	author = {van Zyl, Llewellyn E.},
	year = {2026},
}

@article{ballout_trauma_2025,
	title = {Trauma, {Mental} {Health} {Workforce} {Shortages}, and {Health} {Equity}: {A} {Crisis} in {Public} {Health}},
	journal = {International Journal of Environmental Research and Public Health},
	author = {Ballout, Suha},
	year = {2025},
}

@inproceedings{zhang_forecasting_2025,
	title = {Forecasting {Conversation} {Derailments} {Through} {Generation}},
	booktitle = {Proceedings of {INLG}},
	author = {Zhang, Yunfan and McKeown, Kathleen and Muresan, Smaranda},
	year = {2025},
}
\clearpage
\appendix

\newpage
\section{Counselor Improvement}
\label{appendix:convo_outcome}
 \subsection{Measuring and validating conversational outcomes}
We use an LLM-as-judge to annotate conversation outcomes according to a pre-defined rubric and human-annotated examples, as follows.
We use \texttt{gemma-2-9b-it} \cite{riviere2024gemma2} and the prompt in \autoref{fig:prompt_outcome}.
Given that we often observed very clear indicators of success (e.g., texter gratitude and relief) or unsuccess (e.g., signals of disengagement or continued distress) at the end of conversations, 
we feed the model as input the conversation's last $6$ turns, and ask the model to rate how well the conversation ended on a scale of $1$--$5$ following the pre-defined rubric and examples in the prompt. 

We validate the scores using both automatic and human evaluation, as follows.

For automatic evaluation, we focus on conversations in which the texter disengages, as is essentially signaled by the counselors themselves: they are trained to deploy certain templated check-in language 
(e.g., ``stepped away from your phone,'' ``haven’t heard from you [in a while],'' ``wanted to check in''), when they suspect the texter has stopped replying.
We identify such cases using rule-based heuristics  \cite{nguyen_hanging_2025} and verify LLM scores correlate with this signal (Spearman's $\rho = 0.47, p < 0.0001$; $94$\% of disengaged conversations were assigned a score $\le 3$ by LLM-judge).

For human validation, we construct a pairwise comparison task in which an annotator is presented with two conversations differing by at least $2$ points and asked to identify which ended better.
One author performed this task,\footnote{Due to highly sensitive nature of the data, outside annotators cannot be used.} correctly guessing the label of $47$ out of $50$ pairs, showing high agreement with the LLM-judge ($94$\% accuracy). 
We use this score as our measure of conversational quality, forming the basis for quantifying counselor performance over the course of their careers.

\subsection{Characterizing counselor improvement}
\begin{figure}[t]
    \centering
    \includegraphics[width=.8\linewidth]{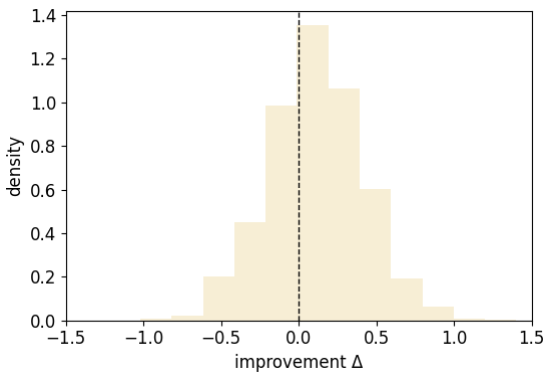}
    \caption{Distribution of counselor improvement.%
    }
    \label{fig:improvement}
\end{figure}

\begin{figure}[t]
    \centering
    \includegraphics[width=.8\linewidth]{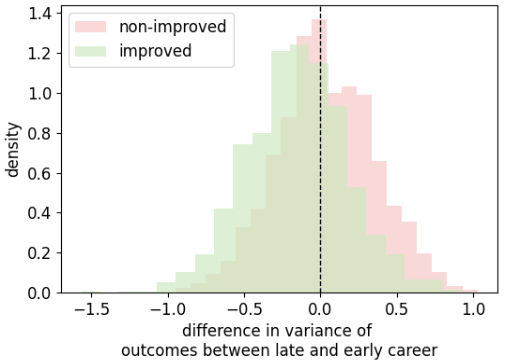}
    \caption{Distribution of the change in variance of outcomes between late and early careers of improved and nonimproved counselors.}
    \label{fig:change_variance}
\end{figure}

In \autoref{sec:improvement}, we introduced our measure of counselor improvement.
To further understand what improvement looks like, \autoref{fig:improvement} shows the distribution of improvement scores across counselors. 
\autoref{fig:change_variance} indicates the change in variance of conversational outcome scores between late and early career of improved vs. nonimproved counselors. %

\begin{figure}[t]
\centering
\small
\fbox{
\begin{minipage}{0.9\linewidth}
You will be given a transcript of the end of a crisis counseling conversation between a mental health counselor and a texter in distress. Your task is to evaluate the outcome of the conversation based on how the conversation ends. Specifically, determine whether at the end of the conversation, the texter feels supported, heard, and reaches a sense of closure, versus disengages, withdraws, or does not show improvement. Assign a score from 1 to 5 reflecting the outcome quality, with lower scores indicating disengagement or poorer outcomes, and higher scores indicating positive resolution and support. \\

Here is the annotation scale: \\
1 – Negative \\
Texter disengages or withdraws from the conversation.
2 – Poor \\
Texter shows minimal improvement with moderate levels of doubt or distress. \\
3 – Neutral / Mixed \\
Texter shows some signs of supported, but slight hesitancy or distress remains. \\
4 – Positive \\
Texter feels moderately supported, understood, calmer, or more hopeful, with minimal hesitancy or distress. \\
5 – Strong Positive \\
Texter clearly demonstrates a clear sense of being supported, understood, calmer, or more hopeful. The conversation ends with a clear sense of closure and satisfaction. If the conversation meets these criteria, assign 5.  \\

Here are some example annotations with scores: \\
\{\{ Example with score 1 \}\} \\
\{\{ Example output and explanation 1 \}\} \\

\{\{ Example with score 2 \}\} \\
\{\{ Example output and explanation 2 \}\} \\

\{\{ Example with score 3 \}\} \\
\{\{ Example output and explanation 3 \}\} \\

\{\{ Example with score 4 \}\} \\
\{\{ Example output and explanation 4 \}\} \\

\{\{ Example with score 5 \}\} \\
\{\{ Example output and explanation 5 \}\} \\

Now, here is the transcript to analyze:
\{\{ transcript \}\} \\ 

Assign a score from 1 (negative) to 5 (positive) reflecting the outcome of the conversation and provide a short explanation for your answer. Use the full 1–5 scale. Do not avoid extreme scores (1 or 5). Be decisive and avoid conservative scoring. Do not default to 4 if 5 clearly applies (see Example 5 again). Base your judgment on the examples above for calibration. Format your answer as a JSON dictionary with the score and explanation. Just output the JSON, nothing else. \\

Your output should follow this format: \\
\{\{
  "score": "score from 1 to 5", \\
  "explanation": "very short explanation for the score" \\
\}\}

Output:
\end{minipage}
}
\caption{Prompt used to label conversation outcomes. }
\label{fig:prompt_outcome}
\end{figure}

\section{Operationalization}
\label{appendix:operationalization}

\begin{table*}[t]
\centering
\begin{tabular}{lcccc}
\toprule
Method & Acc & P & R & F1 \\ 
\midrule
Downturn (unsuccessful) $\rightarrow$ paired similar moment
& \textbf{59.7} & \textbf{59.1} & 63.8 & 61.1 \\
Downturn (all) $\rightarrow$ paired similar moment
& 59.3 & 58.6 & 64.2 & 61.1 \\
Downturn (unsuccessful) $\rightarrow$ paired random moment
& 59.4 & 58.2 & \textbf{67.7} & \textbf{62.3} \\
\midrule
High sentiment drop (unsuccessful) $\rightarrow$ paired similar moment
& 56.8 & 57.0 & 63.0 & 58.7 \\
Random moment (unsuccessful) $\rightarrow$ paired similar moment
& 57.9 & 58.4 & 59.6 & 58.1 \\
Random moment (all) $\rightarrow$ paired similar moment
& 55.2 & 57.7 & 37.3 & 45.0 \\
\bottomrule
\end{tabular}
\caption{Full validation results for \textit{detecting} whether a counselor has improved across different moment selections and pairings (early $\rightarrow$ late). Averaged across multiple seeds.}
\label{tab:detection_val_all}
\end{table*}

\subsection{Downturns}
To pinpoint downturns, we rely on the use of a conversational forecasting model, trained to predict, as a conversation develops, its eventual outcome. 
We follow the same setup as prior work in this domain \cite{nguyen_hanging_2025} which fine-tuned RoBERTa-large \cite{liu_roberta_2019} using a dataset of $5000$ conversations, exactly half  of which end with the texter disengaging. 
We trained for $5$ epochs, with learning rate $1e-5$, batch size
$16$, and AdamW optimizer \cite{loshchilov_decoupled_2017}.
The model achieves a forecasting accuracy of $73$\%, following the evaluation methodology introduced in prior work on conversation forecasting \cite{chang_trouble_2019}.

\subsection{Summarizer and classifier details}
To generate adaptations, we use \texttt{gemma-3-4b-it} \cite{gemmateam2025gemma3} as the base summarizer with the prompt given in  \autoref{fig:prompt_adaptation}.
We set the temperature to $0.8$ and top\_p to $0.9$.
For the classifier, we use \texttt{gemma-3-1b-pt} \cite{gemmateam2025gemma3} as the backbone model. 
We finetune our classifiers using LoRA \cite{hu_lora_2021}, applied to the query and value projection matrices, using $r = 16$, $\alpha=32$, dropout $0.05$, for $3$ epochs, batch size $8$, learning rate $1e-4$, weight decay $0.01$, and context length $N=1024$ for summaries and $N=4096$ for transcripts.
We use lightweight models for the classifier and summarizer due to the computational cost, with the consideration of RL-training as well. 
All work was done on secure internal servers due to the private nature of the data. 

\subsection{RL-training details}
To learn adaptations more indicative of improvement, we jointly train the summarizer and classifier. 
We implement this training setup using OpenRLHF \cite{hu-etal-2025-openrlhf}.
Training the classifier follows the same settings as described above. 
We train the summarizer via GRPO \cite{shao_deepseekmath_2024} using the classifier as the reward signal. 
We train for $1$ epoch for each iteration, with batch size $32$, actor learning rate $5e-7$, critic learning rate $9e-6$, KL coefficient $0.01$,  max length $4096$, $5$ rollouts and AdamW optimizer \cite{loshchilov_decoupled_2017}. 

\subsection{Additional implementation details}
To construct our comparison sets for adaptations, we pair each downturn with the most contextually similar moment later in a counselor's career.
For similarity, we use a lightweight sentence embedding model \texttt{all-MiniLM-L6-v2} for the representation and cosine similarity to compare the contexts.

For the sentiment baseline, we use the \texttt{distilbert-base-uncased-finetuned-sst2} 
\texttt{-english} model.

\subsection{Used artifacts}
We indicate the following artifacts and their corresponding licenses that are used in this work. 

\begin{itemize}[leftmargin=*]
    \item ConvoKit 4.0.0:
    \\ \url{https://convokit.cornell.edu/}, MIT License
    \item PyTorch 2.2.1:
    \\ \url{https://pytorch.org}, BSD-3 License 
    \item Sentence Transformers 3.0.0:
    \\\url{https://github.com/UKPLab/sentence-transformers}, Apache License 2.0
    \item Transformers 4.38.2:
    \\ \url{https://github.com/huggingface/transformers}, Apache License 2.0
\end{itemize}

\begin{figure}[t]
\centering
\small
\fbox{
\begin{minipage}{0.9\linewidth}
You will be given two sets of messages. The messages could be from different conversations and are at different points in the conversation with different people. The messages focus on: all. \\

Now provide a concise summary how the speaker stayed the same vs. changed in Set 2 from Set 1 based on the messages. Focus on the overall similarities and differences in aggregate. Do not based your answer off isolated examples, focus on overall patterns. Only describe patterns that appear multiple times. Write the summary strictly from Set 2's perspective, in terms of how Set 2 is from Set 1. \\

Do not write from Set 1's perspective. The style you should avoid: "While Set 1 ..., Set 2..." \\

Instead say: "Set 2 is more ... and less ..." \\

Each set contain messages from the same speaker across different conversations during an unknown time period. The goal is to summarize how this person changed or not based on their messages in Set 2 compared to Set 1 and the degree to which they changed or not. When generating your answer, think about the messages that correspond to each part, but don't mention them in the output. Remember these are messages that could appear across different conversations in different contexts. Only describe key different patterns that appear multiple times and are the most salient. Do not consider specific topics or content. Avoid interpreting whether the dynamics are effective or not, just analyze them in a neutral tone. Focus on "how" it is said. Your response should be 150-200 words. \\

Please analyze the interactional similarities and differences between the two sets and provide a concise summary of them as a whole.\\ 
\{\{ Set 1 \}\} \\
\{\{ Set 2 \}\} \\

Use simple language and sentences, be broad, and do not include the message text in your answer. Mention the degree of change or no change if there is minimal change in "how" things are said.\\
Output:
\end{minipage}
}
\caption{Prompt used for adaptation summaries.}
\label{fig:prompt_adaptation}
\end{figure}

\begin{table*}[t]
\centering
\begin{tabular}{lcccc}
\toprule
Method & Acc & P & R & F1 \\ 
\midrule
Downturn (unsuccessful) $\rightarrow$ paired similar moment
& \textbf{59.2} & \textbf{58.5} & 65.5 & \textbf{61.4} \\
Downturn (all) $\rightarrow$ paired similar moment
& 56.9 & 56.0 & \textbf{65.6} & 60.1 \\
Downturn (unsuccessful) $\rightarrow$ paired random moment
& 58.9 & 58.1 & 63.3 & 60.4 \\
\midrule
High sentiment drop (unsuccessful) $\rightarrow$ paired similar moment
& 54.6 & 54.4 & 62.2 & 56.8 \\
Random moment (unsuccessful) $\rightarrow$ paired similar moment
& 55.7 & 55.5 & 57.5 & 55.6 \\
Random moment (all) $\rightarrow$ paired similar moment
& 52.2 & 52.9 & 40.6 & 45.7 \\
\midrule
Downturn (unsuccessful) $\rightarrow$ paired similar moment \textbf{with RL}
& 65.1 & 66.4 & 61.0 & 63.6 \\
\bottomrule
\end{tabular}
\caption{Full test results for \textit{detecting} whether a counselor has improved across different moment selections and pairings (early $\rightarrow$ late). Averaged across multiple seeds.}
\label{tab:detection_test_all}
\end{table*}

\section{Additional Results}
\label{appendix:additional-results}

Here, we also show more detailed results for the post-hoc detection task, detecting whether a counselor had improved. 
\autoref{tab:detection_val_all} contains the results on validation, and \autoref{tab:detection_test_all} contains the results on test. 

\begin{figure}[H]

    \centering
    \includegraphics[width=.7\linewidth]{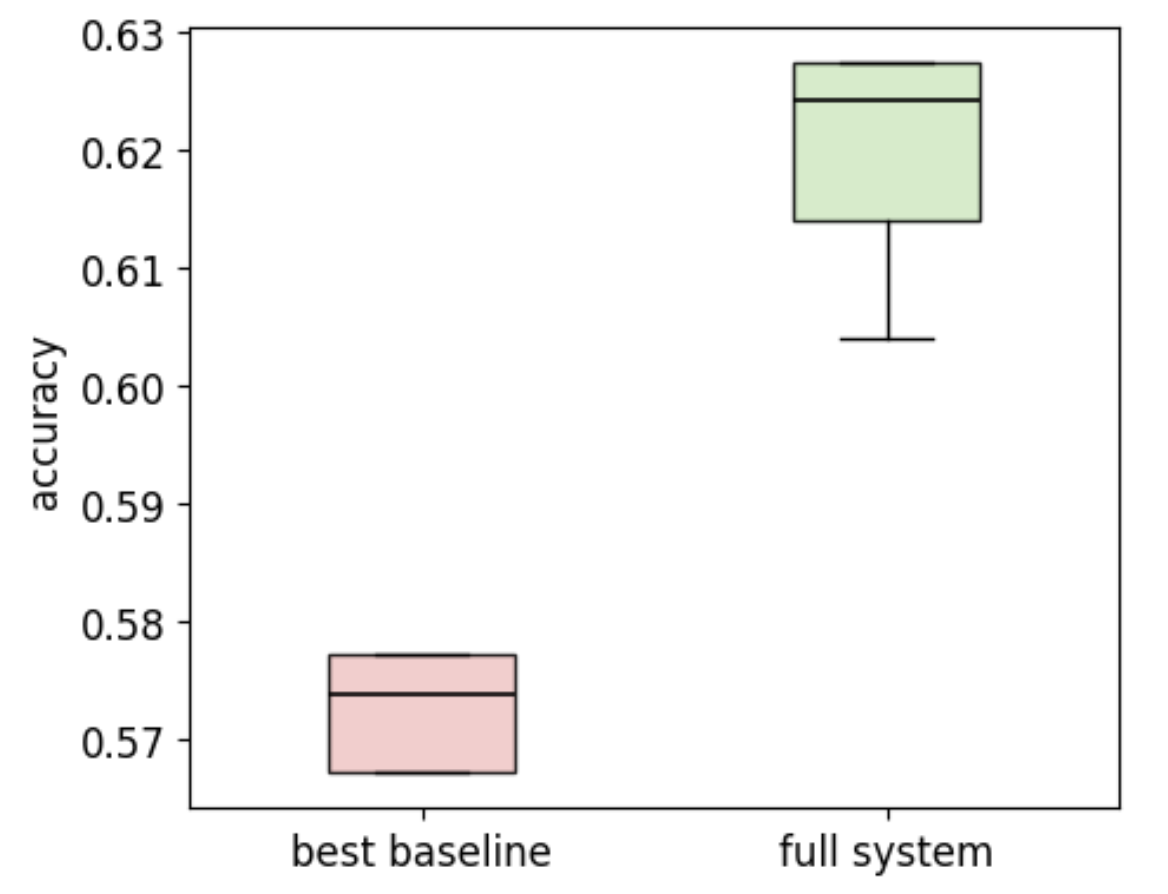}
    \caption{
    Box and whiskers plot comparing the accuracy of our full system (right) and the best baseline: \(\times\) msg. (left). Whiskers not visible due to overlap with box edges.
    }
    \label{fig:seed_results}
\end{figure}

For the early prediction task, we also compare the performance of our full system and the best baseline (\(\times\) msg.) across multiple seeds in \autoref{fig:seed_results}, showing that our system systematically outperforms the best baseline.

\section{Additional Qualitative Examples}
\label{appendix:additional-qual}
\subsection{Types of downturn moments}
We break down downturns into different types via a clustering procedure.
We use KMeans to cluster the embeddings of the immediate texter messages leading up to the downturn. 
We then performed a Bayesian distinguishing word-analysis \cite{monroe_fightin_2017} to compare phrases that distinguish each cluster from the rest. 
Using these distinguishing phrases and examples in each cluster, we roughly label the clusters as follows: (1) \textit{seeking support, clarification} (2) \textit{introduction, self-disclosure} (3) \textit{isolation, hopelessness} (4) \textit{relationships, conflict} (5) \textit{suicidal ideation, self-harm} (6) \textit{anxiety, depression, panic attacks}. 
\autoref{tab:clusters_trigrams} summarizes the distinguishing phrases and \autoref{tab:qual_examples} shows example moments from each cluster (paraphrased for privacy).

\subsection{Counselor downturns}
We also show the distribution of downturns across counselors,
that is, for each downturn type, the proportion of counselors who have that type as the most common one. 
Significant variation is evident.

    \includegraphics[width=.8\linewidth]{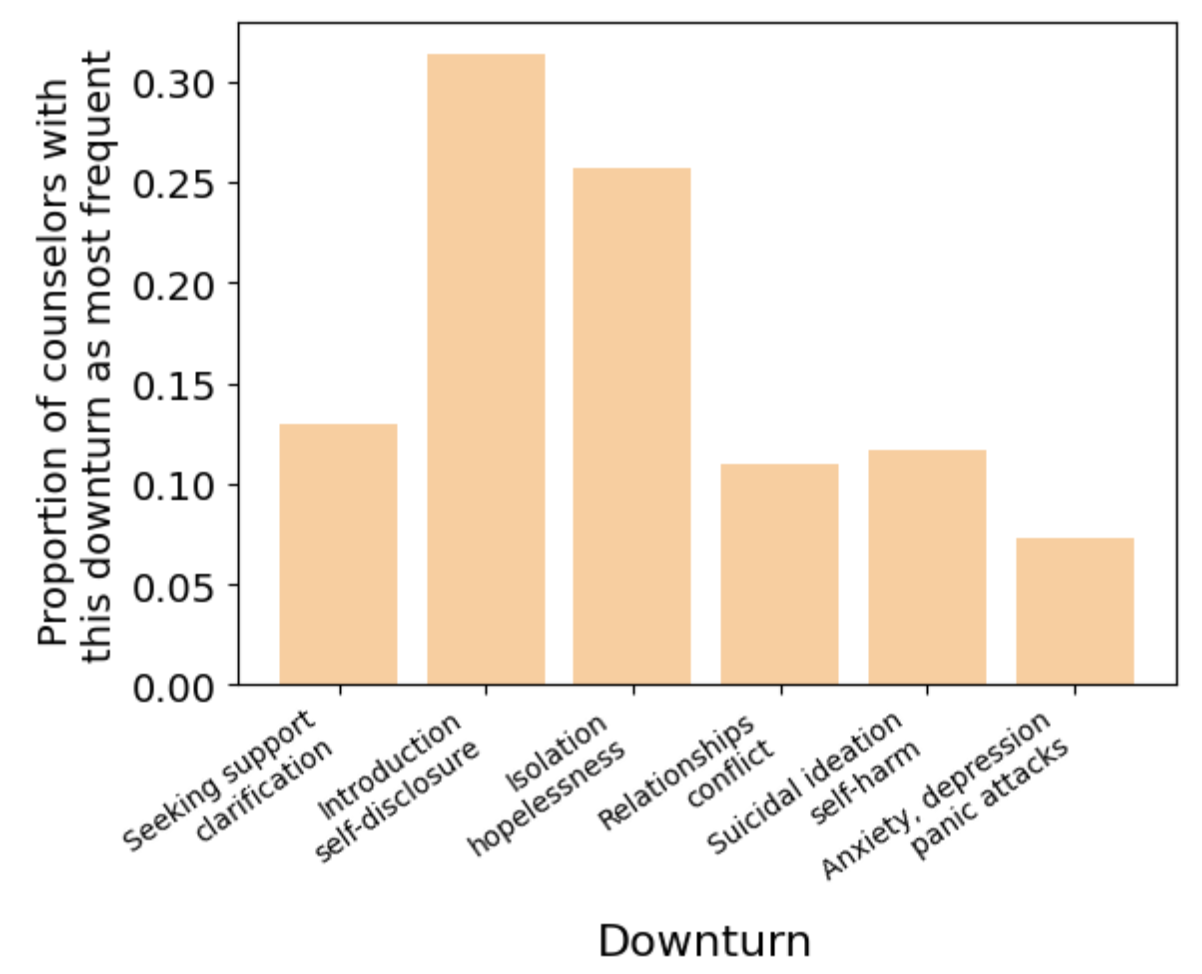}

\subsection{Additional analysis}
We now turn towards understanding how counselors may adapt to these moments over time and how such adaptations may differ between counselors who improved vs.~not. 
Our analysis follows the same examples summarized in \autoref{tab:qual_examples}. 
In general, counselors who improved may also ``shift towards a more encouraging tone,'' ``express gratitude for sharing,'' become ``more focused on validating the [texter's] feelings,'' and prioritize ``reassurance and encouragement over detailed questioning'' (``You're not alone anymore, I have your back...''; Example 2). 
The interactional style may be characterized by a greater degree of empathy (Examples 2, 3).

In contrast, counselors who do not improve can also change in different ways that may not lead to improvement.
In Example 4, a counselor who did not improve became ``considerably more focused on problem-solving,'' with a ``move toward suggesting coping mechanisms'' and adopting a ``more active and interventionist role'' in previous downturns.
A counselor in Example 5 actually seems to be more reassuring in their early career, but later opted for ``more direct and questioning statements,'' a more ``consistently inquisitive tone'' and ``objective communication'' style later on.
Counselors may also become ``more conversational and less formal'' \cite{zhang_finding_2019}.
In this example, the counselor's ``communication style remains consistent in empathy'' as well.

\begin{table*}[h]
\centering
\small
\setlength{\tabcolsep}{6pt}
\begin{tabular}{l r | l r}
\toprule
\multicolumn{2}{c|}{\textbf{Not before downturn }} &
\multicolumn{2}{c}{\textbf{Before downturn}} \\
\cmidrule(lr){1-2} \cmidrule(lr){3-4}
\textbf{n-gram} & \textbf{$z$} &
\textbf{n-gram} & \textbf{$z$} \\
\midrule
have to go             & $-7.07$ & my name is                 & $+10.12$ \\
do you think           & $-5.80$ & name is scrubbed     & $+9.81$ \\
what do you            & $-5.02$ & want to die               & $+6.25$ \\
going to go            & $-4.67$ & and want to               & $+5.46$ \\
no no no               & $-4.45$ & to kill myself            & $+5.45$ \\
do you mean            & $-4.36$ & want to kill              & $+4.98$ \\
but thank you          & $-3.84$ & start want to             & $+4.27$ \\
just try to            & $-3.80$ & need someone to           & $+4.15$ \\
it makes me            & $-3.79$ & my names scrubbed   & $+3.96$ \\
thank you for          & $-3.76$ & hi scrubbed scrubbed & $+3.95$ \\
listen to music        & $-3.64$ & having panic attack       & $+3.85$ \\
not at the             & $-3.57$ & need to talk              & $+3.84$ \\
you for listening      & $-3.46$ & is scrubbed and     & $+3.76$ \\
going to sleep         & $-3.46$ & go want to                & $+3.53$ \\
figure out how         & $-3.39$ & end my life               & $+3.45$ \\
be able to             & $-3.36$ & scrubbed my name    & $+3.43$ \\
to talk anymore        & $-3.32$ & to end my                 & $+3.37$ \\
no don want            & $-3.25$ & going through lot         & $+3.34$ \\
for your time          & $-3.25$ & and have no               & $+3.34$ \\
and they said          & $-3.15$ & to do scrubbed      & $+3.33$ \\
don wanna talk         & $-3.15$ & my friend is              & $+3.24$ \\
going to help          & $-3.07$ & home want to              & $+3.20$ \\
it doesn help          & $-2.99$ & going to kill             & $+3.18$ \\
this isn the           & $-2.99$ & lot going on              & $+3.16$ \\
to go now              & $-2.97$ & like giving up            & $+3.13$ \\
the way my             & $-2.97$ & like killing myself       & $+3.11$ \\
have to come           & $-2.96$ & kill myself scrubbed & $+3.06$ \\
thanks for listening   & $-2.90$ & to die scrubbed     & $+3.06$ \\
got to go              & $-2.86$ & lot of anxiety            & $+3.06$ \\
have good night        & $-2.86$ & need help please          & $+2.98$ \\
\bottomrule
\end{tabular}
\caption{Top 30 tri-grams associated with texter messages that appear right before a downturn (right) or not (left), ranked by the z-score of the Bayesian distinguishing-word analysis.}
\label{tab:random_lowri_ngrams}
\end{table*}

\begin{table*}[t]
\centering
\small
\setlength{\tabcolsep}{4pt}
\begin{tabularx}{\textwidth}{
    >{\raggedright\arraybackslash}p{2.2cm}
    >{\raggedright\arraybackslash}X
}
\toprule
\textbf{Downturn} & \textbf{Distinguishing trigrams} \\
\midrule
Seeking support, clarification &
``yes it is'', ``is this real'', ``this real person'', ``okay thank you'', 
``are you real'', ``do you mean'', ``don think so'', ``yes ma am'', 
``what do you'', ``no not really'', ``do have to'', ``ok thank you'', 
``no don know'', ``thank you for'', ``how do you'', ``would like to'', 
``you real person'', ``at the moment'', ``know my name'', ``old are you'' \\
\midrule
Introduction, self-disclosure &
``name is scrubbed'', ``my name is'', ``scrubbed my name'', 
``hi scrubbed scrubbed'', ``my names scrubbed'', 
``call me scrubbed'', ``is scrubbed and'', ``hi scrubbed my'', 
``hi my name'', ``scrubbed thank you'', ``thank you scrubbed'', 
``can call me'', ``yes my name'', ``to do scrubbed'', 
``talk to scrubbed'', ``you can call'', ``hello my name'', 
``nice to meet'', ``scrubbed scrubbed scrubbed'', 
``my name scrubbed'' \\
\midrule
Isolation, hopelessness &
``have no one'', ``just feel like'', ``feel so alone'', 
``in my life'', ``my life is'', ``like have no'', 
``to be happy'', ``have no friends'', ``feel like have'', 
``like no one'', ``and feel like'', ``no one to'', 
``feel like not'', ``feel like no'', ``just feel so'', 
``to do anymore'', ``don feel like'', ``feel like giving'', 
``feel like my'', ``and have no'' \\
\midrule
Relationships, conflict &
``he told me'', ``he said he'', ``my best friend'', 
``me and my'', ``mad at me'', ``to me and'', 
``he says he'', ``doesn want to'', ``broke up with'', 
``to help him'', ``talk to her'', ``at me and'', 
``he wants to'', ``up with me'', ``and he just'', 
``we ve been'', ``him and he'', ``want him to'', 
``with him and'', ``talk to him'' \\
\midrule
Suicidal ideation, self-harm &
``want to die'', ``to kill myself'', ``want to kill'', 
``don want to'', ``kill my self'', ``going to kill'', 
``want to cut'', ``end my life'', ``to end my'', 
``home want to'', ``to kill my'', ``to be alive'', 
``help want to'', ``to cut myself'', ``start want to'', 
``want to end'', ``hello want to'', ``hate my life'', 
``to commit suicide'', ``want to be'' \\
\midrule
Anxiety, depression, panic attacks &
``having panic attack'', ``an anxiety attack'', ``anxiety and depression'', 
``my anxiety is'', ``depression and anxiety'', ``lot of anxiety'', 
``having an anxiety'', ``go to sleep'', ``having really bad'', 
``really bad anxiety'', ``to calm down'', ``my thoughts are'', 
``my anxiety and'', ``ve been having'', ``panic attack and'', 
``an eating disorder'', ``my mind is'', ``eating disorder and'', 
``during the day'', ``bad anxiety and'' \\
\bottomrule
\end{tabularx}
\caption{Distinguishing trigrams by downturn type.}
\label{tab:clusters_trigrams}
\end{table*}

\begin{table*}[ht]
\centering
\small
\setlength{\tabcolsep}{6pt}
\begin{tabularx}{\textwidth}{
    >{\centering\arraybackslash}p{0.1cm}
    >{\raggedright\arraybackslash}p{1.4cm}
    >{\raggedright\arraybackslash}X
    >{\raggedright\arraybackslash}X
    >{\raggedright\arraybackslash}X
}
\toprule
\textbf{\#} & \textbf{Downturn} & \textbf{Early career response} & \textbf{Late career response} & \textbf{Summary excerpt} \\
\midrule

1 &
Suicidal ideation, self-harm &
\textbf{T:} I’m feeling suicidal \newline
\textbf{C:} Hi, I'm scrubbed. I'm here for you, tell me more. \newline
\textbf{T:} I've been wanting to die for a week and idk what to do \newline
\textbf{C: You're brave for being so open with me tonight. Would you be comfortable telling me your first name?}
&
\textbf{T:} I am suicidal \newline
\textbf{C:} Hi, I hear you're feeling overwhelmed tonight. Can you tell me more? \newline
\textbf{T:} I feel alone and wanna die \newline
\textbf{C: I'm here today to support you. What's happened recently that made you feel this way?}
&
[...] demonstrates a consistent pattern of offering support and expressing concern for the individual’s well-being [...] leans more heavily on direct statements of care [...] the approach is generally more proactive in seeking clarification about the individual’s situation and needs [...] \\
\midrule

2 &
Introduction, self-disclosure &
\textbf{T:} I can’t meet my family’s expectations. It’s too much \newline
\textbf{C:} It makes sense to feel this way. How long has this been?\newline
\textbf{T:} A year, I just need someone to talk to \newline
\textbf{C: I am here to help you through your crisis. You will not always feel this way, it'll get better. Do you mind sharing your name with me?}
&
\textbf{T:} My names scrubbed, my family is too much \newline
\textbf{C:} That must be really tough to have your family shut you down like that. How are you feeling? \newline
\textbf{T:} I feel frustrated and all alone \newline
\textbf{C: Thank you for sharing that with me. You're not alone anymore, I have your back and you can tell me anything if you want?}
&
[...] shift towards a more supportive and encouraging tone [...] more focused on validating the speaker’s feelings and strengths, expressing gratitude for their sharing [...] prioritizing reassurance and encouragement over detailed questioning [...] overall interactional style is characterized by a greater degree of empathy \\
\midrule

3 &
Relationships, conflict &
\textbf{T:} Okay, I'm scrubbed. \newline
\textbf{C:} Thanks, scrubbed. Can you tell me more about the conflict going on at home? \newline
\textbf{T:} My dad is angry and rude towards all the time. I'm afraid to be myself around him because he's so negative. \newline
\textbf{C: That sounds like an awful situation to live with. How long has this been?}
&
\textbf{T:} Yes it’s scrubbed. \newline
\textbf{C:} It sounds like you're under much stress, tell me more? \newline
\textbf{T:} I'm going to attend dream music school. My dad can be abusive if he found out\newline
\textbf{C: You sound like a really creative person, congrats! It must be super stressful w/your dad's reaction, you don't deserve that.}
&
[...] frequently starts with expressions of empathy and concern [...] utilizes phrases of affirmation and encouragement [...] overall approach is less exploratory and more immediately concerned with providing a supportive and protective environment [...] \\
\midrule
\midrule

4 &
Isolation, hopelessness &
\textbf{T:} I used to be really happy but this sadness just hit me \newline
\textbf{C:} I am glad you shared that with me. It seems that you are feeling sad but you don't really know why. \newline
\textbf{T:} Exactly it’s hopeless \newline
\textbf{C: That must be frustrating for you.}
&
\textbf{T:} What’s wrong with me \newline
\textbf{C:} That's not something I can answer, how are you feeling? \newline
\textbf{T:} I've been unusually sad for a long time I can't remember when I was happy  \newline
\textbf{C: Have you considered journaling about your feelings to sort them out?}
&
[...] tone is considerably more focused on problem-solving [...] conversational style is less tentative, with a move toward suggesting coping mechanisms like journaling [...] move toward a more active and interventionist role \\

\midrule

5 &
Suicidal ideation, self-harm &
\textbf{T:} I want to die. \newline
\textbf{C:} Has anything happened to make you want to self-harm? \newline
\textbf{T:} I felt terrible and getting blamed everything \newline
\textbf{C: I am sorry for your loss sweetheart, it can be rather nasty when people unjustly blame you for things. Have you felt like this for a while?} \newline
&
\textbf{T:} I'll be dead soon. I wish I had spared myself years of pain. \newline
\textbf{C:} You're upset you are going to die soon? \newline
\textbf{T:} I am upset that my life ended. I am essentially already dead. \newline
\textbf{C: I am sorry to hear that you are feeling this way, though do you still want to continue talking?}
&
While [early set] frequently used terms of endearment like “sweetheart” and offered reassurance, [late set] opts for more direct and questioning statements. [...] more distanced interaction style [...] indicates a move towards a more objective communication [...]
\\
\midrule

6 &
Isolation, hopelessness &
\textbf{T:} I'm lonely, I have no friends \newline
\textbf{C:} Can you tell me more? \newline
\textbf{T:} I just lost my best friend since she was making me try to change myself. I have nobody. \newline
\textbf{C: That sounds devastating. You deserve to be treated with respect. Are you comfortable sharing your first name with me?}
&
\textbf{T:} I’m kinda lonely \newline
\textbf{C:} What's going on? \newline
\textbf{T:} My girlfriend is away, all my friends are gone so I'm alone. \newline
\textbf{C: It can be hard when the people you rely on aren't around. That can be tough to manage.}
&
[...] acknowledging the speaker’s experience and validating their feelings [...] phrasing is slightly more conversational and less formal [...] the communication style remains consistent in its empathy, but it’s subtly altered by a stronger desire to understand the speaker’s perspective [...] \\

\bottomrule
\end{tabularx}
\caption{
Paraphrased downturn and later analogous moments with adaptation summary excerpts, separated by \textit{improved} counselors (top) and \textit{nonimproved} counselors (bottom). Adaptation summary compares the entire collection of downturns and analogous moments from the same counselor, not just the one exemplified.
}
\label{tab:qual_examples_all}
\end{table*}

\section{AI Disclosure}
\label{appendix:ai}
In this section, we disclose the use of AI. AI was used for spellchecking and grammar correction on a fully written text and for formatting LaTeX tables. AI was also used to help find related papers.

\label{appendix:appendixsection}

\end{document}